\documentclass[letter, 10 pt, conference]{ieeeconf}  

\IEEEoverridecommandlockouts 
\usepackage[english]{babel}
\usepackage[utf8]{inputenc}
\usepackage{times} 
\usepackage{cite}

\usepackage{hyperref}
\usepackage[nolist]{acronym}
\usepackage{footmisc}

\usepackage{multirow}

\usepackage{siunitx}
\usepackage{graphicx} 
\usepackage{epsfig} 
\usepackage{pgfplots}
\usepackage{caption}
\usepackage[font=footnotesize]{subcaption}
\usepackage{pgfplots}

\usepackage{float}

\usepackage[export]{adjustbox}

\newcommand{\white}[1]{
	{\color{white}{#1}}%
}

\usepackage{mathtools}
\usepackage{nicefrac}
\usepackage{amsmath}
\usepackage{amssymb}
\usepackage{bm} 
\usepackage{balance}

\DeclareMathOperator*{\argmin}{arg\,min}

\usepackage{algorithm}
\usepackage[noend]{algpseudocode}

\makeatletter
\def\BState{\State\hskip-\ALG@thistlm}
\makeatother

\usepackage{pgfgantt}
\usepackage{dtsc-creafig}

\usepackage{tikz}
\pgfdeclarelayer{foreground}
\pgfsetlayers{background, main, foreground}
\usetikzlibrary{quotes, angles, backgrounds, arrows, automata, shapes, positioning, calc, through, spy, decorations.pathreplacing, decorations.markings, arrows.meta, automata, petri}

\tikzset{
    imglabel/.style={
      rectangle,
      inner sep=2pt,
      text=black,
      minimum height=1em,
      text centered,
      fill=white,
      fill opacity=1.0,
      text opacity=1,
      anchor=south west,
    },
  }
\tikzset{
	state/.style={
		rectangle,
		draw=black, very thick,
		minimum height=1.0em,
		text centered,
	},
}
\tikzset{
  on each segment/.style={
    decorate,
    decoration={
      show path construction,
      moveto code={},
      lineto code={
        \path [#1]
        (\tikzinputsegmentfirst) -- (\tikzinputsegmentlast);
      },
      curveto code={
        \path [#1] (\tikzinputsegmentfirst)
        .. controls
        (\tikzinputsegmentsupporta) and (\tikzinputsegmentsupportb)
        ..
        (\tikzinputsegmentlast);
      },
      closepath code={
        \path [#1]
        (\tikzinputsegmentfirst) -- (\tikzinputsegmentlast);
      },
    },
  },
  mid arrow/.style={postaction={decorate,decoration={
        markings,
        mark=at position .5 with {\arrow[#1]{stealth}}
      }}},
}

\graphicspath{{./figures/}}

\newcommand\copyrighttext{%
	\small \begin{center} \color{red} \textcopyright\,2022 IEEE. Personal use of this material is permitted. Permission from IEEE must be obtained for all other uses, in any current or future media, including reprinting/republishing this material for advertising or promotional purposes, creating new collective works, for resale or redistribution to servers or lists, or reuse of any copyrighted component of this work in other works. \end{center}}
\newcommand\copyrightnotice{%
	\begin{tikzpicture}[remember picture,overlay]
	\node[anchor=south,yshift=25.6cm] at (current page.south) 
	{\color{red}\fbox{\parbox{\dimexpr\textwidth-\fboxsep-\fboxrule\relax}{\copyrighttext}}};
	\end{tikzpicture}%
}

\title{\copyrightnotice \LARGE \bf Mission Planning and Execution in Heterogeneous Teams of\\ Aerial Robots supporting Power Line Inspection Operations} 

\author{{\'{A}lvaro} Calvo$^1$, Giuseppe Silano$^2$, and Jes{\'{u}}s Capit{\'{a}}n$^1$  
    \thanks{This work is funded by the the European Union's Horizon 2020 research and innovation programme AERIAL-CORE under grant agreement no. 871479.}
    \thanks{$^{1}${\'{A}}lvaro Calvo and Jes{\'{u}}s Capit{\'{a}}n are with the GRVC Robotics Laboratory, University of Seville, 41092 Seville, Spain, (email: {\tt\small \{acalvo1, jcapitan\}@us.es).} }
    \thanks{$^2$Giuseppe Silano is with the Faculty of Electrical Engineering, Department of Cybernetics, Czech Technical University in Prague, 12135 Prague, Czech Republic, (email: {\tt\small giuseppe.silano@fel.cvut.cz).}}
}

\begin{document}
\maketitle
\thispagestyle{empty}
\pagestyle{empty}


\begin{acronym}
    \acro{CBBA}[CBBA]{Consensus-Based Bundle Algorithm}
    \acro{CVRP}[CVRP]{Capacity Vehicle Routing Problem}
    \acro{FCFS}[FCFS]{First-Come-First-Serve}
	\acro{LTL}[LTL]{Linear Temporal Logic}
	\acro{MILP}[MILP]{Mixed-Integer Linear Program}
	\acro{MRS}[MRS]{Multi-Robot System}
	\acro{ROS}[ROS]{Robot Operating System}
	\acro{SIL}[SIL]{Software-in-the-loop}
	\acro{UAV}[UAV]{Unmanned Aerial Vehicle}
	\acro{UV}[UV]{Ultra Violet}
	\acro{TL}[TL]{Temporal Logic}
	\acro{TSP}[TSP]{Traveling Salesman Problem}
	\acro{VRP}[VRP]{Vehicle Routing Problem}
	\acro{wrt}[w.r.t.]{with respect to}
	\acro{BT}[BT]{Behavior Tree}
	\acro{FSM}[FSM]{Finite State Machine}
\end{acronym}


\begin{abstract}

A software architecture aimed at coordinating a team of heterogeneous aerial vehicles for inspection and maintenance operations in high-voltage power line scenarios is presented in this paper. A hierarchical approach deals with high-level tasks by planning and executing complex missions requiring vehicles to support human operators. 
A resource-constrained problem allows distributing tasks among the team taking into account vehicles' capabilities and battery constraints. Besides,~\acp{BT} are in charge of mission execution, triggering replanning operations in case of unforeseen events, such as vehicle faults or communication drop-outs. The feasibility and validity of the approach are showcased through realistic simulations achieved in Gazebo.

\end{abstract}





\section{Introduction}
\label{sec:introduction}


Energy demand has increased significantly over the last decades. In order to keep up with this pace, inspection and maintenance operations in electric power lines and related infrastructures are becoming of uppermost importance for supply companies, as a way to prevent power outages and mitigate economic losses. Nowadays, these operations are usually scheduled periodically and carried out by experienced working crews, who gather inspection data and repair/replace the damaged parts on active lines using manned helicopters. However, this procedure is highly risky, as humans need to operate at height, under windy conditions, and in hazardous environments. 

\begin{figure}[tb]
    \centering
    \begin{subfigure}{0.45\columnwidth}
        \centering
        \adjincludegraphics[width=0.97\textwidth, trim={{0.37\width} {0.3\height} {0.35\width} {0.3\height}}, clip]{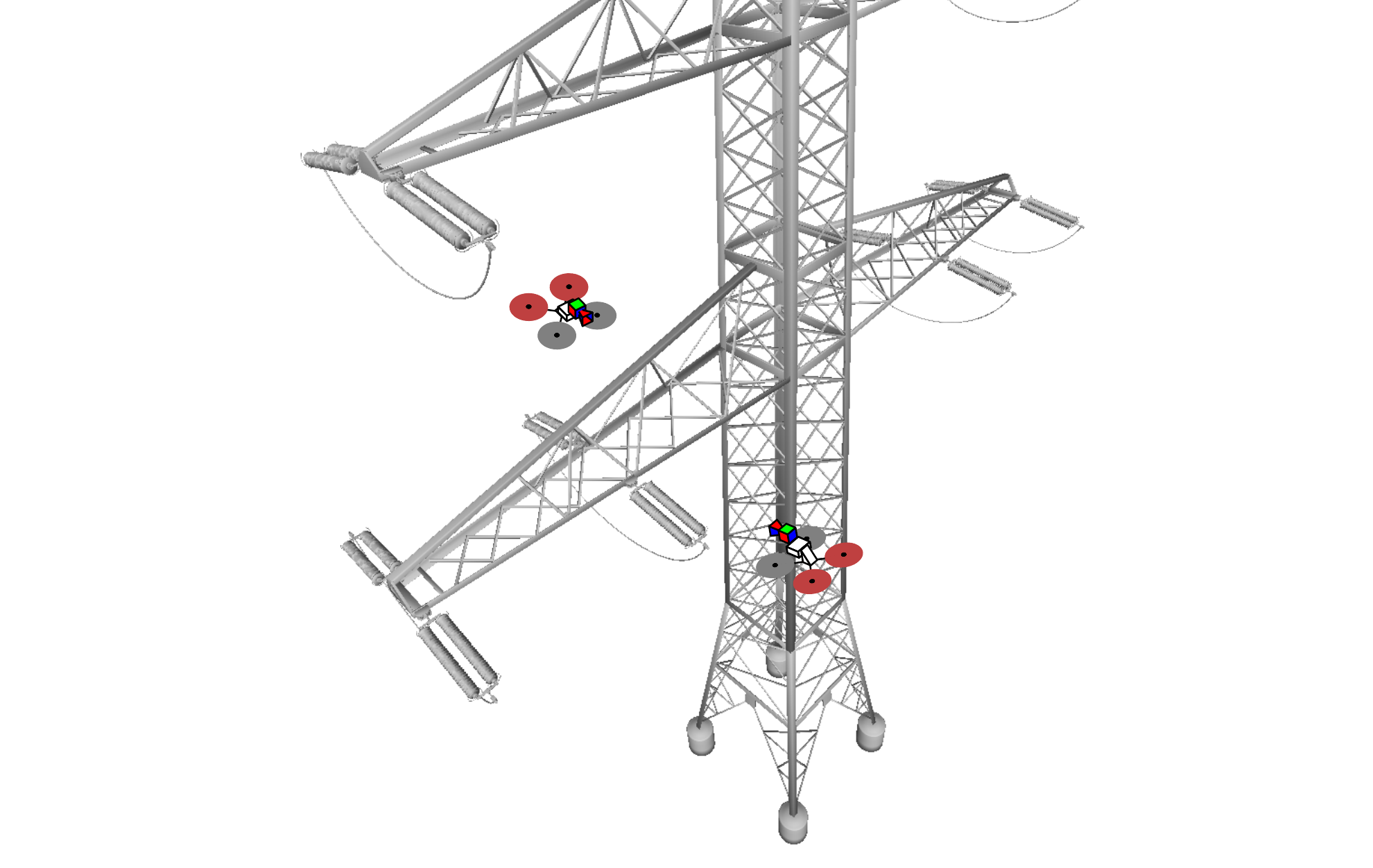}
        \caption{}
        \label{subfig:inspection_task}
    \end{subfigure}
    \hspace{0.5mm}
    \begin{subfigure}{0.45\columnwidth}
        \centering
        \adjincludegraphics[width=1.05\textwidth, trim={{0.0\width} {0.1\height} {0.0\width} {0.0\height}}, clip]{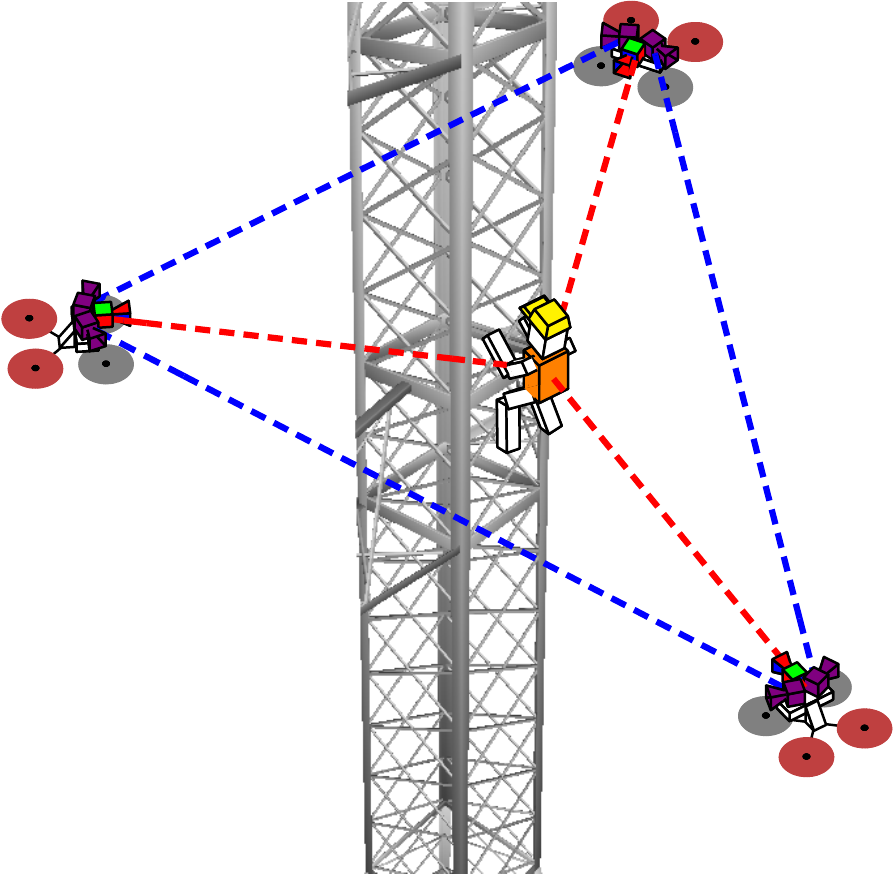}
        \caption{}
        \label{subfig:monitoring_task}
    \end{subfigure}
    \\
    \begin{subfigure}{0.45\columnwidth}
        \centering
        \adjincludegraphics[width=0.95\textwidth, trim={{0.22\width} {0.2\height} {0.0\width} {0.15\height}}, clip]{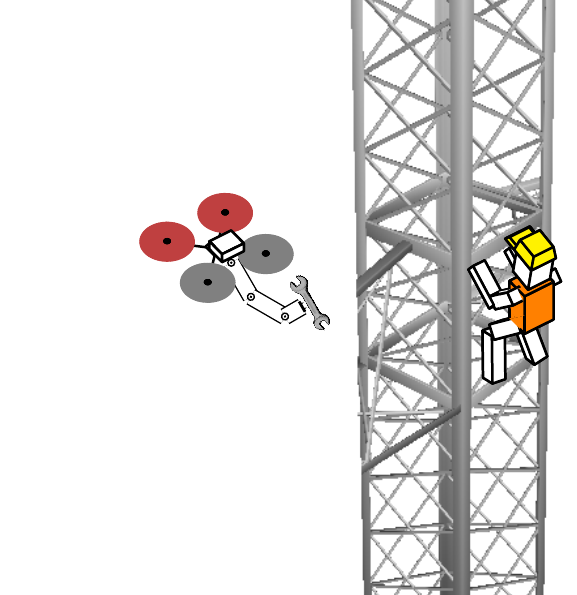}
        \caption{}
        \label{subfig:tool_delivery_task}
    \end{subfigure}
    \vspace{-1mm}
    \caption{Heterogeneous \acp{UAV} supporting inspection and maintenance operations. From left to right: \textit{inspection}, \textit{monitoring}, and \textit{delivery} scenarios.} 
    \label{fig:TaskTypes}
    \vspace{-1em}
\end{figure}

Therefore, there is a strong interest of power suppliers in finding appropriate technologies~\cite{Martinez2018EAAI} to increase safety and efficiency in these maintenance activities.~\acp{UAV} constitute a promising solution to automate inspection operations~\cite{ParkJFR2019}, as they can work in hazardous places, inspect remote locations of difficult access, and monitor human operation for safety purposes. The idea of using a team of heterogeneous~\acp{UAV} cooperating and performing various tasks to support a human crew on site is even more appealing (see Fig.~\ref{fig:TaskTypes}). However, this problem is challenging for several reasons: (i)~\acp{UAV} have limited time of flight and payload, which forces them to schedule recharging operations for a longer endurance; (ii) heterogeneity prohibits the arbitrary exchange of one~\ac{UAV} for another, which makes coordination for proper mission accomplishment more complex; and (iii) the inspection needs to be performed in dynamic settings, where~\acp{UAV} or communication could fail. Therefore, there is a need for cooperative multi-\ac{UAV} architectures where heterogeneous vehicles can efficiently compute and execute plans for long inspection missions, extending their flight autonomy and reacting online to unforeseen events, such as communication drop-outs or vehicle's breakdowns or faults.


In this paper, we propose a software architecture for planning and execution of inspection and maintenance missions with a heterogeneous fleet of multi-rotor~\acp{UAV} to support human crews in power lines assessment operations. In terms of planning, the required tasks are allocated to different~\acp{UAV} depending on their capabilities, and recharging tasks are scheduled for longer operation. Given the dynamic and uncertain conditions of the environment where the mission takes place (e.g, the number of available~\acp{UAV}, the pending tasks, actual battery consumption, tasks' duration, etc.), the execution of the mission is steadily monitored in order to react to unplanned events by recomputing the task assignment online. The software architecture extends and complements our prior work~\cite{SilanoICUAS21}, where we integrated the~\ac{UAV}'s low-level capabilities taking the task planning part for granted.



\subsection{Related work}
\label{sec:relatedWork}

In a multi-robot context, \emph{mission planning} consists of deciding which tasks to allocate to each robot and then building plans with the assignment. \emph{Mission execution} carries out those plans and monitors them for mission accomplishment. Both problems have been widely studied in the literature over years~\cite{choudhury_ar22, korsah_ijrr13} and they can be mainly grouped into two classes, depending on whether the task allocation problem is addressed in a centralized or decentralized fashion. Centralized solutions, in the form of constrained optimization problems, allow to retrieve the best schedule for each robot~\cite{Leahy2021, Miloradovic2021TC}. 
However, these solutions suffer from the high computational burden required to solve the optimization problem. On the other hand, decentralized solutions~\cite{Hou2021, Kim2020} can address the computational load by sharing the problem complexity over the robots. However, such an approach is formulated at the expense of an increase in the inter-robot communication or state estimation, which may be unfeasible in some applications. Therefore, centralized approaches can be more suitable for scenarios with a bounded number of robots, mainly if communication network reliability and compliance with safety requirements are among the mission objectives.

Centralized auction algorithms~\cite{Gerkey2002TRO} and consensus-based methods~\cite{Zhang2020} are commonplace. 
Some of these works present solutions addressing heterogeneity in the robots' capacities~\cite{Leahy2021, Zhang2020}. Whereas, in missions where the tasks are placed at different locations to be visited, routing problem formulations are preferred. However, the latter easily ends up being combinatorial NP-hard problems, so heuristic techniques are usually applied to retrieve the solution within a reasonable time~\cite{dorling_17,agarwal_icra20}. 
Alternatives consider the inclusion of temporal constraints and uncertainties in the tasks' descriptions~\cite{choudhury_ar22}. In this regard, 
\acl{TL}~\cite{Silano2021ICRARAL} 
can deal with the tasks allocation when complex mission requirements need to be met. However, these problems easily become very complex, being non-convex \textit{min-max} optimization problems~\cite{Pola2019ARC}. On the other hand, some works propose multi-\ac{UAV} task allocation frameworks integrating mission planning and execution~\cite{Arbanas2018, alcantara_access20,cashmore_icaps15}. In~\cite{Arbanas2018}, heterogeneous teams with ground and aerial vehicles are combined so that~\acp{UAV} can land by carrying out recharging operations and thus extending their autonomy. In~\cite{alcantara_access20}, event-triggered replanning is considered during the mission execution. ROSPlan~\cite{cashmore_icaps15} focuses on deterministic complex planning problems.

Finally, regarding the encoding of the~\ac{UAV} behavior during mission execution,~\acp{FSM} are still the most widespread option~\cite{real_access21}, although recently the use of Colored Petri nets~\cite{Xiaojun2018ACC} have been also proposed. 
Among these,~\acfp{BT} are gaining momentum~\cite{Colledanchise2017TRO}, thanks to the advantages in terms of behavior composability and reusability, fault tolerance, and parallel task execution.

\subsection{Contributions}
\label{sec:contributions}


In this paper, we propose a software architecture for mission planning and execution of inspection and maintenance tasks with a heterogeneous team of~\acp{UAV} for power lines assessment missions. Our approach is based on a centralized \textit{High-Level Planner}, that assigns tasks to the~\acp{UAV} and schedules recharging operations in between based on operator's requests, along with a distributed \textit{Agent Behavior Manager}, in charge of mission execution on board each vehicle. Specifically, in Section~\ref{sec:problem_description} we define an application-driven problem for supporting human crews during inspection missions. While, in Section~\ref{sec:system_description} we propose a software architecture for multi-\ac{UAV} mission planning and execution. The software stack is released as open-source\footnote{\label{footnote:softwareGitHub}\url{https://github.com/grvcTeam/aerialcore_planning}} making it possible to go though any part of the framework and replicate the obtained results. Our main contributions are as follows:

\begin{itemize}

    \item We propose a heuristic planner (Section~\ref{subsec:high_level_planner}) that can cope with~\acp{UAV} heterogeneous capabilities and battery constraints, allocating tasks and scheduling recharging operations in between.
    
    \item A distributed management component based on~\acp{BT} (Section~\ref{subsec:agent_behavior_manager}) is in charge of monitoring the tasks execution and requesting an online replanning in case of system's breakdowns or failures (e.g., a~\ac{UAV} runs out of battery).
    
    \item Software-In-The-Loop (SITL) simulations have been carried out in realistic simulation scenarios showcasing the performance and feasibility of our methods (Section~\ref{sec:experimental_results}) and providing insight into future directions (Section~\ref{sec:conclusions}). Benefits in terms of computation time for planing missions are also provided.

\end{itemize}



\section{Problem Description}
\label{sec:problem_description}

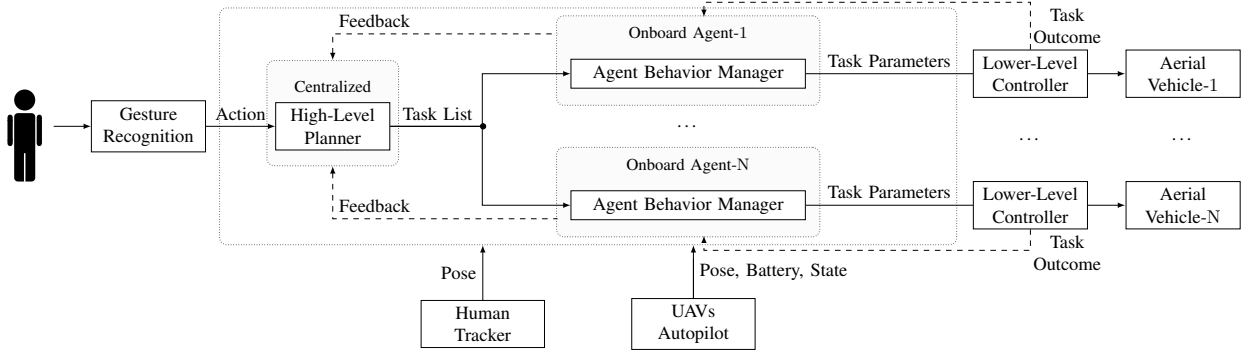
\begin{figure*}[tb]
    \centering
	\scalebox{0.70}{
		\begin{tikzpicture}
    		\node (WP7-Box) at (8.35,0) [rounded corners, densely dotted, minimum height=5cm, minimum width=20cm]{};  

    		\node (TaskPlannerBox) at ($(WP7-Box)+(0,0)$) [fill=white, rounded corners, draw=black!70, densely dotted, minimum height=4.5cm, minimum width=14cm]{}; 
    		
    		\node (GestureRecognition) at (0,0) [text centered, fill=white, draw, rectangle, minimum width=1.5cm, text width=5.5em]{Gesture\\Recognition};
    		
			\node at (-2.35,0.5) [circle,fill,minimum size=4mm] (head) {};
			\node[rounded corners=2pt,minimum height=1.3cm,minimum width=0.4cm,fill,below = 1pt of head] (body) {};
			\draw[line width=1mm,round cap-round cap] ([shift={(2pt,-1pt)}]body.north east) --++(-90:6mm);
			\draw[line width=1mm,round cap-round cap] ([shift={(-2pt,-1pt)}]body.north 
			west)--++(-90:6mm);
			\draw[thick,white,-round cap] (body.south) --++(90:5.5mm);
    		
    		\draw[-latex] ($(GestureRecognition) - (1.8,0)$) -- (GestureRecognition);
 
    		\node (HighLevelPlannerBox) at ($(GestureRecognition) + (3.5,0.25)$) [fill=gray!3, rounded corners, draw=black!70, densely dotted, minimum height=2cm, minimum width=2.5cm]{}; 
    		\node (HighLevelPlanner) at ($(HighLevelPlannerBox) + (0,-0.25)$) [text centered, fill=white, draw, rectangle, minimum width=1.5cm, text width=5.5em]{High-Level\\Planner};
    		\node (Centralized) at ($(HighLevelPlanner) + (0,0.75)$) [text centered]{\small{Centralized}};
    		
    		\draw[-latex] (GestureRecognition.east) -- node[above]{Action} (HighLevelPlanner);
    		
    		\draw ($ (HighLevelPlanner.east) + (0.9,0) $) node[above, text centered]{Task List};
    		
    		\node (UAV1) at ($(HighLevelPlanner) + (6.75,1.25)$) [fill=gray!3, rounded corners, draw=black!70, densely dotted, minimum height=1.7cm, minimum width=5cm]{}; 
    		\node (AgentBehaviorManager1) at ($(UAV1) + (0,-0.25)$) [fill=white, draw, rectangle, text centered, text width=12em]{Agent Behavior Manager};
    		\node (UAV1-Text) at ($(AgentBehaviorManager1) + (0,0.75)$) [text centered]{\small{Onboard Agent-$1$}};	

    		\draw[fill=black] ($ (HighLevelPlanner.east) + (1.715,0) $) arc(-180:180:0.05);
    		\draw[-latex] (HighLevelPlanner.east) -- ($ (HighLevelPlanner.east) + (1.75,0) $) -- ($ (HighLevelPlanner.east) + (1.75,1) $) -- (AgentBehaviorManager1.west);
    		
    		\draw[dashed, latex-] ($ (HighLevelPlanner.north) + (0,0.8) $) -- ($ (HighLevelPlanner.north) + (0,1.3) $) -- ($ (HighLevelPlanner.north) + (4.27,1.3) $);
    		\draw ( $(HighLevelPlanner.north) + (0,1.55) $) node[right]{Feedback};
    		
    		\node (Dots2) at ($(UAV1) + (0,-1.25)$) [text centered]{\dots};
    		
    		\node (UAVN) at ($(HighLevelPlanner) + (6.75,-1.25)$) [fill=gray!3, rounded corners, draw=black!70, densely dotted, minimum height=1.7cm, minimum width=5cm]{}; 
    		\node (AgentBehaviorManagerN) at ($(UAVN) + (0,-0.25)$) [fill=white, draw, rectangle, text centered, text width=12em]{Agent Behavior Manager};
    		\node (UAVN-Text) at ($(AgentBehaviorManagerN) + (0,0.75)$) [text centered]{\small{Onboard Agent-$\mathrm{N}$}};	

    		\draw[-latex] (HighLevelPlanner.east) -- ($ (HighLevelPlanner.east) + (1.75,0) $) -- ($ (HighLevelPlanner.east) + (1.75,-1.5) $) -- (AgentBehaviorManagerN.west);
    		
    		\draw[dashed, latex-] ($ (HighLevelPlanner.south) + (0,-0.3) $) -- ($ (HighLevelPlanner.south) + (0,-1.3) $) -- ($ (HighLevelPlanner.south) + (4.27,-1.3) $);
    	    \draw ( $(HighLevelPlanner.south) + (0,-1.05) $) node[right]{Feedback};
    		
    		\node (LowerLevelControllers1) at ($(AgentBehaviorManager1) + (6.5,0)$) [text centered, fill=white, draw, rectangle, minimum width=1.5cm, text width=5.5em]{Lower-Level\\Controller};
    		
    		\node (Dots3) at ($(LowerLevelControllers1) + (0,-1.25)$) [text centered]{\dots};
    		
    		\node (LowerLevelControllersN) at ($(AgentBehaviorManagerN) + (6.5,0)$) [text centered, fill=white, draw, rectangle, minimum width=1.5cm, text width=5.5em]{Lower-Level\\Controller};
    		
    		\draw[-latex] (LowerLevelControllers1.east) -- ($(LowerLevelControllers1) + (1.8,0)$);
    		\draw[-latex] (LowerLevelControllersN.east) -- ($(LowerLevelControllersN) + (1.8,0)$);
    		
    		
    		\draw (AgentBehaviorManager1.east) -- node[above]{Task Parameters}	(LowerLevelControllers1.west);
    	    
    	    \draw[-latex, dashed] (LowerLevelControllersN.south) --  ( $(LowerLevelControllersN.south) + (0,-0.4) $) -- ($(LowerLevelControllersN.south) + (-6.2,-0.4) $) -- ($(LowerLevelControllersN.south) + (-6.2,-0.12) $);
    	    \draw ($(LowerLevelControllersN.south) + (0.7,-0.9) $) node[above, text width=8em, text centered]{Task\\Outcome};
    	    
    	    \draw[-latex, dashed] (LowerLevelControllers1.north) --  ($(LowerLevelControllers1.north) + (0,0.9) $) -- ($(LowerLevelControllers1.north) + (-6.2,0.9) $) -- ($(LowerLevelControllers1.north) + (-6.2,0.62) $);
    	    \draw ($(LowerLevelControllers1.north) + (0.7,0) $) node[above, text width=8em, text centered]{Task\\Outcome};
    	    
    		\draw (AgentBehaviorManagerN.east) -- node[above]{Task Parameters} (LowerLevelControllersN.west);
    		
    		
    		\node (RealUAVs) at ($(WP7-Box.south) + (2,-1.25)$) [text centered, fill=white, draw, rectangle, minimum width=1.5cm, text width=6em]{UAVs\\Autopilot};
    		\node (Humans) at ($(WP7-Box.south) + (-2,-1.25)$) [text centered, fill=white, draw, rectangle, minimum width=1.5cm, text width=6em]{Human\\Tracker};
    		
    		\draw[-latex] (RealUAVs.north) -- node[right]{Pose, Battery, State} ($(WP7-Box.south) + (2,0.25)$);
    		\draw ($(WP7-Box.south) + (2,0)$) -- (RealUAVs.north);
    		\draw[-latex] (Humans.north) -- node[left]{Pose} ($(WP7-Box.south) + (-2,0.25)$);
    		
    		\node (AerialVehicle1) at ($(LowerLevelControllers1) + (2.9,0)$) [text centered, fill=white, draw, rectangle, minimum width=1.5cm, text width=5.5em]{Aerial\\Vehicle-$1$};
    		
    		\node (Dots4) at ($(AerialVehicle1) + (0,-1.25)$) [text centered]{\dots};
    		
    		\node (AerialVehicleN) at ($(LowerLevelControllersN) + (2.9,0)$) [text centered, fill=white, draw, rectangle, minimum width=1.5cm, text width=5.5em]{Aerial\\Vehicle-$\mathrm{N}$};

	    \end{tikzpicture}
	}
	\caption{The software architecture. Arrows represent the data exchange among blocks, while dotted boxes are used to identify the two layers composing the architecture.}
	\label{fig:NodeDiagram}
	\vspace{-1.5em}
\end{figure*}

Three tasks of interest are considered: (i) \textit{inspection}, where a fleet of~\acp{UAV} carries out a detailed investigation of power equipment on its own, assisting human operators in acquiring views of the power tower that are not easily accessible, as depicted in Fig.~\ref{subfig:inspection_task}; (ii) \textit{monitoring}, where a formation of~\acp{UAV} provides to the supervising team a view of the humans working on the power tower to monitor their status and ensure their safety, as shown in Fig.~\ref{subfig:monitoring_task}; and (iii) \textit{delivery}, where a~\ac{UAV} equipped with loading capabilities interacts with a human worker to deliver a tool, as reported in Fig.~\ref{subfig:tool_delivery_task}. Some tasks only require a single~\ac{UAV}, while others involve several~\acp{UAV} who must work together to achieve a common objective. Moreover, depending on their capabilities, some~\acp{UAV} can perform different types of tasks, by serving as both an inspection and monitoring~\ac{UAV}, whereas those with capabilities for physical interaction are the only ones that can perform delivery tasks.

We assume that the~\acp{UAV} operate in a known environment represented by a previously acquired map, including the position of the power towers and lines. Besides, precise algorithms for~\ac{UAV} localization and navigation are taken for granted. Specifically, the~\acp{UAV} in the team are endowed with a proper set of hardware equipment and \textit{low-level controllers} for the execution of the defined tasks. Moreover, the~\acp{UAV} can detect human gestures~\cite{papaioannidis2021EUSIPCO} that are used to provide high-level actions, such as requests for new tasks or new parameters for a previously requested task.


In the \textit{inspection} operations, the low-level controller encodes the mission execution as a trajectory planning problem where the vehicles need to move from an initial position and through a sequence of target points, while avoiding obstacles and maintaining a safety distance between them~\cite{Silano2021ICRARAL}. In the \textit{monitoring} operations,
 the corresponding low-level controller solves a formation control problem where the vehicles need to keep the human worker within the camera frame during the entire operation, providing complementary views from multiple angles~\cite{Kratky2021RAL, Perez2022ICUAS}. In both tasks, quadrotor~\acp{UAV} support maintenance operations. Last, in \textit{delivery} missions the~\ac{UAV} is equipped with hardware to transport and deliver tools to a human worker. 
The corresponding low-level controller implements human-aware motion planning and control algorithms~\cite{Afifi2022ICRA} for a safe interaction with the worker.

Given these settings, the objective is to coordinate the heterogeneous fleet of~\acp{UAV} to perform all the operator's actions while complying with drones' battery constraints and unforeseen events that may occur during the mission execution, such as the arrival or a failure of an action and a~\ac{UAV} running out of battery or communication drop-outs. 



\section{Software Architecture}
\label{sec:system_description}

The software architecture is organized into two layers: the \textit{High-Level Planner} (Section~\ref{subsec:high_level_planner}) and the \textit{Agent Behavior Manager} (Section~\ref{subsec:agent_behavior_manager}). The former is placed at a ground station, while the latter run onboard the~\acp{UAV}. Figure~\ref{fig:NodeDiagram} describes the overall software architecture. 
The \textit{High-Level Planner} interacts with the human crew through a \textit{Gesture Recognition} block~\cite{papaioannidis2021EUSIPCO}. Such a block encodes the human gestures into actions for the fleet of aerial vehicles. Hence, the \textit{High-Level Planner} outputs a set of tasks each vehicle has to execute to fulfill the mission objectives. Then, $\mathrm{N}$-instances of the \textit{Agent Behavior Manager} take care of extracting the necessary parameters from the assigned tasks (e.g., the waypoints to inspect, the human worker to monitor or to whom deliver a tool, etc.) and coordinate the mission execution based on the inputs provided by the \textit{Human Tracker}~\cite{papaioannidis2021EUSIPCO} (which provides the worker position) and \textit{\ac{UAV} Autopilot} blocks~\cite{real_ijars20}. Continuous \textit{Feedback} from each \textit{Agent Behavior Manager} block, namely the status of the~\ac{BT} and the drone's battery level, is sent back to the planner, so that it can react to certain events. The \textit{Task Outcome} of each task (i.e., success or failure) is also communicated to the \textit{Agent Behavior Manager} by the \textit{Low-Level Controllers}. The chain ends with the control signals generated by the \textit{Low-Level Controller} blocks to make the~\acp{UAV} fly. 



\subsection{High-Level Planner}
\label{subsec:high_level_planner}

The \textit{High-Level Planner} is a centralized module of the software architecture. This module is in charge of tasks planning, i.e., it decides which task will be assigned to each \ac{UAV}. Mission high-level actions (i.e., inspection, monitoring, and delivery) are encoded as human gestures provided by the human crew on the site. Multiple actions could be requested simultaneously. For instance, the crew may ask to inspect a set of target points (i.e., points of interest for the action), performing a visual examination of the power equipment and their surroundings, while the operation of a worker on a nearby power tower is monitored, e.g., providing views with multiple cameras in formation. Recall that there exist multiple~\acp{UAV} for concurrent mission objectives, and that certain vehicles may play different roles (inspection or monitoring) depending on the needs. 

To come up with an action assignment that satisfies the mission objectives, a~\acl{FCFS} scheduling policy is implemented to arrange actions such that maximum priority operations are taken. Besides, actions are endowed with positive numerical weights that can be tuned to parametrize the execution based on the~\ac{UAV} type. Before allocating them to \acp{UAV}, actions are arranged within a queue that follows an ascending order based on the weights that have been assigned to each action. The so-formulated scheduling policy allows us to capture user preferences, i.e., the importance or priorities of different actions, and to adapt the mission in case of incompatible tasks or with performance preferences, by changing the actions' position within the queue.

Once an action has been taken (inspection, monitoring, or delivery), a task allocation problem is formulated to decide which task  (i.e., points of interest for the action) will be assigned to each~\ac{UAV}. A resource-constrained problem deals with the task assignment, where the resources are the number of available \acp{UAV}, their capabilities and remaining battery levels. 
Given the complexity of the problem, which may be intractable when the number of tasks, \acp{UAV} and mission constraints increases, we propose an heuristic to determine adequate feasible solutions. 
Specifically, a minimization~\eqref{eq:minimizationProblem} is carried out over a cost function defined as the sum of three terms: 
\begin{equation}\label{eq:minimizationProblem}
    i^\star = \argmin_{\mathcal{V}}{ J_1(\mathcal{V}) + J_2(\mathcal{V}) + J_3(\mathcal{V})} ,
\end{equation}
where $\mathcal{V}$ refers to the set of available~\acp{UAV} and $i^\star$ represents the optimum tasks sequence for the set of~\acp{UAV} $\mathcal{V}$.
$J_1(\mathcal{V})$ represents a cost per type of~\ac{UAV}, in a way to reward the use of most suitable~\acp{UAV} to cope with the task execution (inspection and monitoring~\acp{UAV} can play the same role). $J_2(\mathcal{V})$ is a travel cost, which takes into consideration the distance separating the~\ac{UAV} position to the task start. Last, $J_3(\mathcal{V})$ is an interruption cost, which penalizes interrupting another task in execution to take a new one, according to the action priorities defined by the user. This way, tasks are assigned to \acp{UAV} with the lowest cost per task. 

This task assignment procedure can be run both when mission planning is conducted offline, and during the execution of the mission, when a replan request may be triggered due to unforeseen or external events. In these situations, the \textit{High-Level Planner} checks whether task reallocation is needed. In particular, replanning is triggered every time a running task is finished, a new task arrives or the parameters of a previously commanded task are modified. Also, replanning is required when a~\ac{UAV} gets disconnected or re-connected due to a communication drop-out or a battery fault occurs, that causes a sudden decrease of the~\ac{UAV} remaining flight time making unfeasible for the vehicle to finish its plan. A watchdog timer helps to manage brief~\ac{UAV} disconnection issues by avoiding replanning operations when not needed.

In order to comply with battery constraints, recharging operations are included as additional tasks in the final plan. Battery consumption is estimated (using a constant rate model) considering the required travel distance for each assigned task. Then, depending on the remaining battery level, recharges are placed either before starting a new task or in the middle of a task, breaking it in two parts, so that the \ac{UAV} can resume later (e.g., inspecting part of the points in an inspection action, stopping to recharge, and then continue). \textit{Artificial} actions so that a \ac{UAV} \textit{waits} on its recharging spot can also be scheduled to synchronize task execution in cases of multi-\ac{UAV} tasks (e.g., if several~\acp{UAV} need to start a monitoring action together). These actions are in addition to those provided by the problem description.

\begin{figure}
    \centering
    \scalebox{0.4}{
        \begin{tikzpicture} 
            \begin{ganttchart}[vgrid={*{22}{dotted}, *{9}{dotted}, *{8}{dotted}}, inline, bar label font=\Large, bar height=.7, title label font=\Huge, title height=1]{1}{40}           \gantttitle{Mission Plan}{40}
                \ganttnewline
                \ganttbar{UAV-1}{1}{40}
                \ganttnewline
                \ganttbar[bar/.append style={draw=red, fill=red!5}]{Delivery}{1}{9}
                \ganttbar[bar/.append style={draw=green, fill=green!5}]{Inspect}{10}{17}
                \ganttbar[bar/.append style={draw=gray, fill=gray!5}]{Wait}{18}{22}
                \ganttbar[bar/.append style={draw=blue, fill=blue!5}]{Monitoring}{23}{32}
                \ganttnewline
                \ganttbar{UAV-2}{1}{40}
                \ganttnewline
                \ganttbar[bar/.append style={draw=green, fill=green!5}]{Inspect}{1}{13}
                \ganttbar[bar/.append style={draw=blue, fill=blue!5}]{Monitoring}{14}{22}
                \ganttbar[bar/.append style={draw=orange, fill=orange!5}]{Recharge}{23}{32}
                \ganttbar[bar/.append style={draw=blue, fill=blue!5}]{Monitoring}{33}{40}
            \end{ganttchart}
        \end{tikzpicture}}
    \caption{Example of a mission plan with 2~\acp{UAV} and 4 actions. Different colors indicate the type of action, while their duration is represented by the horizontal axis.}
    \label{fig:recharging_plans}
\end{figure}
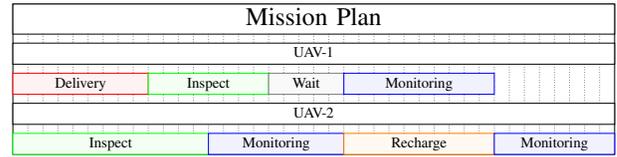

Figure~\ref{fig:recharging_plans} shows an illustrative example with the plans for $2$~\acp{UAV} performing a mission considering $4$ actions. The human worker supplies the~\ac{UAV} team with a list of actions in the order: delivery a tool, inspect an area of interest, perform a monitoring operation, and then inspect another part. For the considered scenario, only the~\ac{UAV}-1 is supposed to be capable of performing delivery tool operations. Given the limited battery capacity,~\ac{UAV}-2 breaks its \textit{monitoring} operation to recharge in the middle. Meanwhile,~\ac{UAV}-1 takes its role. An artificial \textit{wait} action is scheduled before for task synchronization. 

After mission planning, the \textit{High-Level Planner} outputs a list of assigned and feasible tasks for each \ac{UAV}. The tasks are provided as inputs to the $\mathrm{N}$-instances of the \textit{Agent Behavior Manager}, as many as the available vehicles, which are in charge of extracting the tasks' parameters for the corresponding \textit{Low-Level Controllers} where complex behaviors are encoded~\cite{Silano2021ICRARAL, Kratky2021RAL, Perez2022ICUAS, Afifi2022ICRA}.

%



\subsection{Agent Behavior Manager}
\label{subsec:agent_behavior_manager}

The \textit{Agent Behavior Manager} runs as a distributed control chain on board each~\ac{UAV}. This module mainly implements a state machine encoded as a~\ac{BT}, which monitors the~\ac{UAV} state and the task outcome and reacts to any possible failure or unexpected events, activating contingency actions and requesting a new plan to the \textit{High-Level Planner}. Such an approach preserves the key properties of the low-level control systems while ensuring high modularity of the whole control structure. It is indeed worth mentioning that~\acp{BT} unlike~\acp{FSM} guarantee bidirectional control transfers (up and down the control chain) by replacing the concept of \textit{transitions} with the propagation of a signal through the tree. A complete description of the~\ac{BT} syntax, semantics and functioning can be found in~\cite{Colledanchise2017TRO}, and here it is not reported for the sake of brevity.

In short, let a~\ac{BT} be a directed tree embedding the usual definition of nodes, root, leaves, children, and parents. In a~\ac{BT} each node belongs to one of the following categories: \textit{Fallback}, \textit{Sequence}, \textit{Parallel}, \textit{Action}, and \textit{Condition}. Succeeds, Fails, and Running conditions of these node types are summarized in Table~\ref{tab:conditionsBT}. Leaf nodes are either \textit{Actions} or \textit{Conditions}, while interior nodes are either \textit{Fallbacks}, \textit{Sequences}, or \textit{Parallels}. Each leaf represents a particular conclusion or action to be carried out, and each nonleaf represents a predicate to be checked.~\acp{BT} operate propagating a \textit{tick} signal from the root downwards, checking nodes' status according to their operating rules, until it reaches a leaf node, and executing each node's callback in the process.

\begin{table}
	\centering
	\begin{adjustbox}{max width=0.95\columnwidth}
		\begin{tabular}{l|l|l|l}
		    \hline
			\textsc{\textbf{Node type}} & \textsc{\textbf{Succeeds}} & \textsc{\textbf{Fails}} & \textsc{\textbf{Running}}
			\\ 
			\hline \hline
			\shortstack[l]{\textit{Fallback}\\ \white{a}} & \shortstack[l]{If one child\\succeeds} & \shortstack[l]{If all\\children fail}  & \shortstack[l]{If one child\\returns running} \\
			\shortstack[l]{\textit{Sequence}\\ \white{a}} & \shortstack[l]{If all children\\succeed} & \shortstack[l]{If one\\child fails} & \shortstack[l]{If one child\\returns running} \\
			\shortstack[l]{\textit{Parallel} \\ \white{a}} & \shortstack[l]{If $\alpha \in \mathbb{N}^+$\\ children succeed} & \shortstack[l]{If $\beta > \alpha$ children\\fail, with $\beta \in \mathbb{N}^+$} & \shortstack[l]{Else \\ \white{a}} \\
			\shortstack[l]{\textit{Action} \\ \white{a}} & \shortstack[l]{Upon\\completion} & \shortstack[l]{When impossible\\to complete} & \shortstack[l]{During\\completion} \\
			\textit{Condition} & If true & If false & Never \\
			\hline
		\end{tabular}
	\end{adjustbox}
	\caption{Nodes type of a~\ac{BT}.}
	\label{tab:conditionsBT}
\end{table}

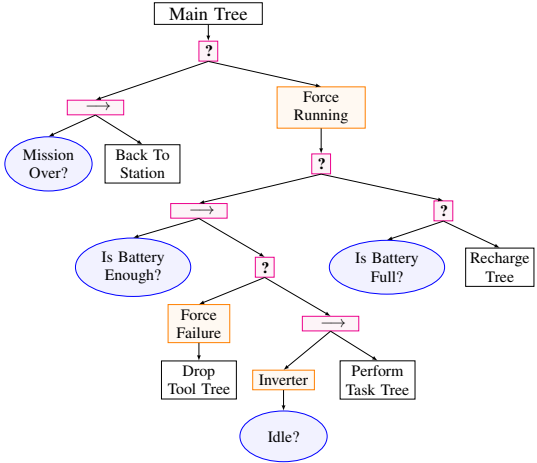
\begin{figure}
	\begin{center}
		\scalebox{0.5}{
			\begin{tikzpicture}
        		\node (MainTree) at (0,0) [text centered, fill=white, draw, rectangle, minimum width=1.5cm, text width=7.5em, font=\Large]{Main Tree};

        		\node (RootFallback) at ($(MainTree) + (0,-1)$) [text centered, fill=magenta!5, draw=magenta, rectangle, minimum width=0.5cm, text width=0.5em, font=\large]{\textbf{?}};
        		\draw[-latex] (MainTree.south) -- (RootFallback.north);

        		\node (MissionOverSequence) at ($(RootFallback) + (-3,-1.5)$) [text centered, fill=magenta!5, draw=magenta, rectangle, minimum width=1.5cm, text width=1.5em, font=\large]{$\longrightarrow$};
        		\draw[-latex] (RootFallback.south) -- (MissionOverSequence.north);
        		\node (ForceRunning) at ($(RootFallback) + (3, -1.5)$) [text centered, fill=orange!5, draw=orange, rectangle, minimum width=1.5cm, text width=6em, font=\large]{Force Running};
        		\draw[-latex] (RootFallback.south) -- (ForceRunning.north);
        		
        		\node (MissionOver) at ($(MissionOverSequence) + (-1.25,-1.5)$) [text centered, fill=blue!5, draw=blue, ellipse, minimum width=1.5cm, text width=4em, font=\large]{Mission Over?};
        		\draw[-latex] (MissionOverSequence.south) -- (MissionOver.north);
        		\node (BackToStation) at ($(MissionOverSequence) + (1.25, -1.5)$) [text centered, fill=white, draw, rectangle, minimum width=1.5cm, text width=5em, font=\large]{Back To Station};
        		\draw[-latex] (MissionOverSequence.south) -- (BackToStation.north);
        		\node (MissionFallback) at ($(ForceRunning) + (0,-1.5)$) [text centered, fill=magenta!5, draw=magenta, rectangle, minimum width=0.5cm, text width=0.5em, font=\large]{\textbf{?}};
        		\draw[-latex] (ForceRunning.south) -- (MissionFallback.north);

        		\node (IsBatteryEnoughSequence) at ($(MissionFallback) + (-3.25,-1.25)$) [text centered, fill=magenta!5, draw=magenta, rectangle, minimum width=1.5cm, text width=1.5em, font=\large]{$\longrightarrow$};
        		\draw[-latex] (MissionFallback.south) -- (IsBatteryEnoughSequence.north);
        		\node (WaitFallback) at ($(MissionFallback) + (3.25,-1.25)$) [text centered, fill=magenta!5, draw=magenta, rectangle, minimum width=0.5cm, text width=0.5em, font=\large]{\textbf{?}};
        		\draw[-latex] (MissionFallback.south) -- (WaitFallback.north);
        		
        		\node (IsBatteryEnough) at ($(IsBatteryEnoughSequence) + (-1.75,-1.5)$) [text centered, fill=blue!5, draw=blue, ellipse, minimum width=1.5cm, text width=5.5em, font=\large]{Is Battery Enough?};
        		\draw[-latex] (IsBatteryEnoughSequence.south) -- (IsBatteryEnough.north);
        		\node (DropToolFallback) at ($(IsBatteryEnoughSequence) + (1.75,-1.5)$) [text centered, fill=magenta!5, draw=magenta, rectangle, minimum width=0.5cm, text width=0.5em, font=\large]{\textbf{?}};
        		\draw[-latex] (IsBatteryEnoughSequence.south) -- (DropToolFallback.north);
        		\node (IsBatteryFull) at ($(WaitFallback) + (-1.5,-1.5)$) [text centered, fill=blue!5, draw=blue, ellipse, minimum width=1.5cm, text width=5.5em, font=\large]{Is Battery Full?};
        		\draw[-latex] (WaitFallback.south) -- (IsBatteryFull.north);
        		\node (RechargeTree) at ($(WaitFallback) + (1.5, -1.5)$) [text centered, fill=white, draw, rectangle, minimum width=1.5cm, text width=4.5em, font=\large]{Recharge Tree};
        		\draw[-latex] (WaitFallback.south) -- (RechargeTree.north);
        		
        		\node (ForceFailure) at ($(DropToolFallback) + (-1.75, -1.5)$) [text centered, fill=orange!5, draw=orange, rectangle, minimum width=1.5cm, text width=4em, font=\large]{Force Failure};
        		\draw[-latex] (DropToolFallback.south) -- (ForceFailure.north);
        		\node (TaskSequence) at ($(DropToolFallback) + (1.75,-1.5)$) [text centered, fill=magenta!5, draw=magenta, rectangle, minimum width=1.5cm, text width=1.5em, font=\large]{$\longrightarrow$};
        		\draw[-latex] (DropToolFallback.south) -- (TaskSequence.north);

        		\node (DropToolTree) at ($(ForceFailure) + (0, -1.5)$) [text centered, fill=white, draw, rectangle, minimum width=1.5cm, text width=5em, font=\large]{Drop Tool Tree};
        		\draw[-latex] (ForceFailure.south) -- (DropToolTree.north);
        		\node (Inverter) at ($(TaskSequence) + (-1.25, -1.5)$) [text centered, fill=orange!5, draw=orange, rectangle, minimum width=1.5cm, text width=4em, font=\large]{Inverter};
        		\draw[-latex] (TaskSequence.south) -- (Inverter.north);
        		\node (PerformTaskTree) at ($(TaskSequence) + (1.25, -1.5)$) [text centered, fill=white, draw, rectangle, minimum width=1.5cm, text width=5em, font=\large]{Perform Task Tree};
        		\draw[-latex] (TaskSequence.south) -- (PerformTaskTree.north);
        		
        		\node (Idle) at ($(Inverter) + (0,-1.5)$) [text centered, fill=blue!5, draw=blue, ellipse, minimum width=1.5cm, minimum height=1.35cm, text width=4em, font=\large]{Idle?};
        		\draw[-latex] (Inverter.south) -- (Idle.north);
        		
		    \end{tikzpicture}}
		\caption{Main Tree. Reactive control nodes (e.g., \textit{Fallback} or \textit{Sequence}) are shown in magenta, leaf nodes (\textit{Action} and \textit{Condition}) in blue, and decorator nodes (to transform the result of a child) in orange. Normal control nodes and sub-trees are in white. }
		\label{fig:MainTree}
	\end{center}
	\vspace{-1em}
\end{figure}

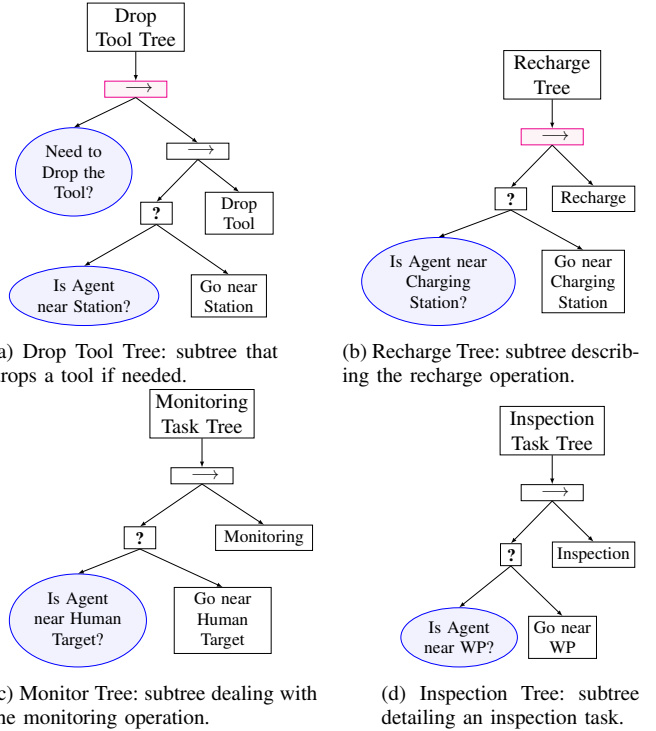
\begin{figure}
    \centering
    \subfloat[Drop Tool Tree: subtree that drops a tool if needed.]{
        \centering
        \scalebox{0.55}{
            \begin{tikzpicture}
                \node (DropToolTree) at (0,0) [text centered, fill=white, draw, rectangle, minimum width=1.5cm, text width=6em, font=\Large]{Drop Tool Tree};
        		
        		\node (ReactiveSequence) at ($(DropToolTree) + (0,-1.5)$) [text centered, fill=magenta!5, draw=magenta, rectangle, minimum width=1.5cm, text width=1.5em, font=\large]{$\longrightarrow$};
        		\draw[-latex] (DropToolTree.south) -- (ReactiveSequence.north);
        		
        		\node (NeedToDropTheTool) at ($(ReactiveSequence) + (-1.5,-2)$) [text centered, fill=blue!5, draw=blue, ellipse, minimum width=1.5cm, text width=5em, font=\large]{Need to Drop the Tool?};
        		\draw[-latex] (ReactiveSequence.south) -- (NeedToDropTheTool.north);
        		\node (Sequence) at ($(ReactiveSequence) + (1.5,-1.5)$) [text centered, fill=white, draw, rectangle, minimum width=1.5cm, text width=1.5em, font=\large]{$\longrightarrow$};
        		\draw[-latex] (ReactiveSequence.south) -- (Sequence.north);
        		
        		\node (Fallback) at ($(Sequence) + (-1,-1.5)$) [text centered, fill=white, draw, rectangle, minimum width=0.5cm, text width=1.5em, font=\large]{\textbf{?}};
        		\draw[-latex] (Sequence.south) -- (Fallback.north);
        		\node (DropTool) at ($(Sequence) + (1,-1.5)$) [text centered, fill=white, draw, rectangle, minimum width=1.5cm, text width=4em, font=\large]{Drop Tool};
        		\draw[-latex] (Sequence.south) -- (DropTool.north);
        		
        		\node (IsAgentNearStation) at ($(Fallback) + (-1.75,-2)$) [text centered, fill=blue!5, draw=blue, ellipse, minimum width=1.5cm, text width=6.6em, font=\large]{Is Agent near Station?};
        		\draw[-latex] (Fallback.south) -- (IsAgentNearStation.north);
        		\node (GoNearStation) at ($(Fallback) + (1.75, -2)$) [text centered, fill=white, draw, rectangle, minimum width=1.5cm, text width=4.5em, font=\large]{Go near Station};
        		\draw[-latex] (Fallback.south) -- (GoNearStation.north);
            \end{tikzpicture}}
	    \label{subfig:DropToolTree}}
    \hfill
    \subfloat[Recharge Tree: subtree describing the recharge operation.]{
        \centering
        \scalebox{0.55}{
            \begin{tikzpicture}
                \node (RechargeTree) at (0,0) [text centered, fill=white, draw, rectangle, minimum width=1.5cm, text width=6em, font=\Large]{Recharge Tree};
        		
        		\node (RechargeTaskSequence) at ($(RechargeTree) + (0,-1.5)$) [text centered, fill=magenta!5, draw=magenta, rectangle, minimum width=1.5cm, text width=1.5em, font=\large]{$\longrightarrow$};
        		\draw[-latex] (RechargeTree.south) -- (RechargeTaskSequence.north);
        		
        		\node (NearChargingStationFallback) at ($(RechargeTaskSequence) + (-1,-1.5)$) [text centered, fill=white, draw, rectangle, minimum width=0.5cm, text width=1.5em, font=\large]{\textbf{?}};
        		\draw[-latex] (RechargeTaskSequence.south) -- (NearChargingStationFallback.north);
        		\node (Recharge) at ($(RechargeTaskSequence) + (1,-1.5)$) [text centered, fill=white, draw, rectangle, minimum width=1.5cm, text width=5em, font=\large]{Recharge};
        		\draw[-latex] (RechargeTaskSequence.south) -- (Recharge.north);
        		
        		\node (IsAgentNearChargingStation) at ($(NearChargingStationFallback) + (-1.75,-2)$) [text centered, fill=blue!5, draw=blue, ellipse, minimum width=1.5cm, text width=6.8em, font=\large]{Is Agent near Charging Station?};
        		\draw[-latex] (NearChargingStationFallback.south) -- (IsAgentNearChargingStation.north);
        		\node (GoNearHuman) at ($(NearChargingStationFallback) + (1.75, -2)$) [text centered, fill=white, draw, rectangle, minimum width=1.5cm, text width=5em, font=\large]{Go near Charging Station};
        		\draw[-latex] (NearChargingStationFallback.south) -- (GoNearHuman.north);
            \end{tikzpicture}}
	    \label{subfig:RechargeTaskTree}}
    \newline
    \subfloat[Monitor Tree: subtree dealing with the monitoring operation.]{
        \centering
		\scalebox{0.55}{
			\begin{tikzpicture}
			    \node (MonitorTree) at (0,0) [text centered, fill=white, draw, rectangle, minimum width=1.5cm, text width=6.5em, font=\Large]{Monitoring Task Tree};
        		
        		\node (MonitorTaskSequence) at ($(MonitorTree) + (0,-1.5)$) [text centered, fill=white, draw, rectangle, minimum width=1.5cm, text width=1.5em, font=\large]{$\longrightarrow$};
        		\draw[-latex] (MonitorTree.south) -- (MonitorTaskSequence.north);
        		
        		\node (NearHumanFallback) at ($(MonitorTaskSequence) + (-1.5,-1.5)$) [text centered, fill=white, draw, rectangle, minimum width=0.5cm, text width=1.5em, font=\large]{\textbf{?}};
        		\draw[-latex] (MonitorTaskSequence.south) -- (NearHumanFallback.north);
        		\node (MonitorHumanTarget) at ($(MonitorTaskSequence) + (1.5,-1.5)$) [text centered, fill=white, draw, rectangle, minimum width=1.5cm, text width=6em, font=\large]{Monitoring};
        		\draw[-latex] (MonitorTaskSequence.south) -- (MonitorHumanTarget.north);
        		
        		\node (IsUAVnearHumanTarget) at ($(NearHumanFallback) + (-1.5,-2)$) [text centered, fill=blue!5, draw=blue, ellipse, minimum width=1.5cm, text width=6em, font=\large]{Is Agent near Human Target?};
        		\draw[-latex] (NearHumanFallback.south) -- (IsUAVnearHumanTarget.north);
        		\node (GoNearHuman) at ($(NearHumanFallback) + (2, -2)$) [text centered, fill=white, draw, rectangle, minimum width=1.5cm, text width=6em, font=\large]{Go near Human Target};
        		\draw[-latex] (NearHumanFallback.south) -- (GoNearHuman.north);
		    \end{tikzpicture}}
		\label{subfig:MonitorTree}}
    \hfill
    \subfloat[Inspection Tree: subtree detailing an inspection task.]{
        \centering
		\scalebox{0.55}{
			\begin{tikzpicture}
			    \node (InspectTree) at (0,0) [text centered, fill=white, draw, rectangle, minimum width=1.5cm, text width=6.5em, font=\Large]{Inspection Task Tree};
        		
        		\node (InspectTaskSequence) at ($(InspectTree) + (0,-1.5)$) [text centered, fill=white, draw, rectangle, minimum width=1.5cm, text width=1.5em, font=\large]{$\longrightarrow$};
        		\draw[-latex] (InspectTree.south) -- (InspectTaskSequence.north);
        		
        		\node (NearWPFallback) at ($(InspectTaskSequence) + (-1,-1.5)$) [text centered, fill=white, draw, rectangle, minimum width=0.5cm, text width=0.5em, font=\large]{\textbf{?}};
        		\draw[-latex] (InspectTaskSequence.south) -- (NearWPFallback.north);
        		\node (Inspect) at ($(InspectTaskSequence) + (1,-1.5)$) [text centered, fill=white, draw, rectangle, minimum width=1.5cm, text width=5em, font=\large]{Inspection};
        		\draw[-latex] (InspectTaskSequence.south) -- (Inspect.north);
        		
        		\node (IsUAVnearWP) at ($(NearWPFallback) + (-1.25,-2)$) [text centered, fill=blue!5, draw=blue, ellipse, minimum width=1.5cm, text width=5em, font=\large]{Is Agent near WP?};
        		\draw[-latex] (NearWPFallback.south) -- (IsUAVnearWP.north);
        		\node (GoNearWP) at ($(NearWPFallback) + (1.25, -2)$) [text centered, fill=white, draw, rectangle, minimum width=1.5cm, text width=4em, font=\large]{Go near WP};
        		\draw[-latex] (NearWPFallback.south) -- (GoNearWP.north);
		    \end{tikzpicture}}
		\label{subfig:InspectTree}}
    \caption{Subtrees encoding the basic actions for the tasks described in Section~\ref{sec:problem_description}.}
    \label{fig:someBTs}
\end{figure}

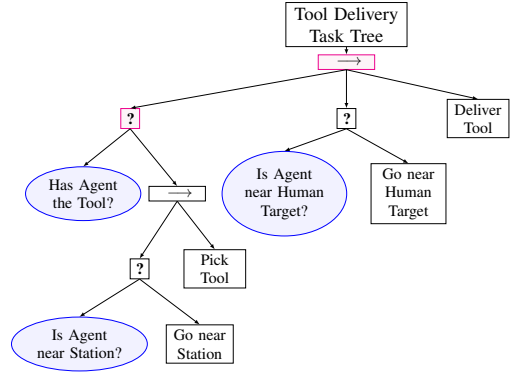
\begin{figure}
	\begin{center}
		\scalebox{0.5}{
			\begin{tikzpicture}
			    \node (DeliverTree) at (0,0) [text centered, fill=white, draw, rectangle, minimum width=1.5cm, text width=8.5em, font=\Large]{Tool Delivery Task Tree};
			    
			    \node (DeliverTaskSequence) at ($(DeliverTree) + (0,-1)$) [text centered, fill=magenta!5, draw=magenta, rectangle, minimum width=1.5cm, text width=1.5em, font=\large]{$\longrightarrow$};
        		\draw[-latex] (DeliverTree.south) -- (DeliverTaskSequence.north);

				\node (ToolFallback) at ($(DeliverTaskSequence) + (-5.75,-1.5)$) [text centered, fill=magenta!5, draw=magenta, rectangle, minimum width=0.5cm, text width=0.5em, font=\large]{\textbf{?}};
        		\draw[-latex] (DeliverTaskSequence.south) -- (ToolFallback.north);
        		\node (HumanFallback) at ($(DeliverTaskSequence) + (0,-1.5)$) [text centered, fill=white, draw, rectangle, minimum width=0.5cm, text width=0.5em, font=\large]{\textbf{?}};
        		\draw[-latex] (DeliverTaskSequence.south) -- (HumanFallback.north);
				\node (DeliverTool) at ($(DeliverTaskSequence) + (3.5,-1.5)$) [text centered, fill=white, draw, rectangle, minimum width=1.5cm, text width=4em, font=\large]{Deliver Tool};
        		\draw[-latex] (DeliverTaskSequence.south) -- (DeliverTool.north);

				\node (hasACWtheTool) at ($(ToolFallback) + (-1.25,-2)$) [text centered, fill=blue!5, draw=blue, ellipse, minimum width=1.5cm, text width=5.5em, font=\large]{Has Agent the Tool?};
        		\draw[-latex] (ToolFallback.south) -- (hasACWtheTool.north);
        		\node (PickToolSequence) at ($(ToolFallback) + (1.25,-2)$) [text centered, fill=white, draw, rectangle, minimum width=1.5cm, text width=1.5em, font=\large]{$\longrightarrow$};
        		\draw[-latex] (ToolFallback.south) -- (PickToolSequence.north);
        		\node (IsUAVnearHuman) at ($(HumanFallback) + (-1.65,-2)$) [text centered, fill=blue!5, draw=blue, ellipse, minimum width=1.5cm, text width=6em, font=\large]{Is Agent near Human Target?};
        		\draw[-latex] (HumanFallback.south) -- (IsUAVnearHuman.north);
        		\node (GoNearHuman) at ($(HumanFallback) + (1.65, -2)$) [text centered, fill=white, draw, rectangle, minimum width=1.5cm, text width=5em, font=\large]{Go near Human Target};
        		\draw[-latex] (HumanFallback.south) -- (GoNearHuman.north);

        		\node (StationFallback) at ($(PickToolSequence) + (-1,-2)$) [text centered, fill=white, draw, rectangle, minimum width=0.5cm, text width=0.5em, font=\large]{\textbf{?}};
        		\draw[-latex] (PickToolSequence.south) -- (StationFallback.north);
        		\node (PickTool) at ($(PickToolSequence) + (1, -2)$) [text centered, fill=white, draw, rectangle, minimum width=1.5cm, text width=4em, font=\large]{Pick Tool};
        		\draw[-latex] (PickToolSequence.south) -- (PickTool.north);
        		
        		\node (IsUAVnearStation) at ($(StationFallback) + (-1.6,-2)$) [text centered, fill=blue!5, draw=blue, ellipse, minimum width=1.5cm, text width=6.6em, font=\large]{Is Agent near Station?};
        		\draw[-latex] (StationFallback.south) -- (IsUAVnearStation.north);
        		\node (GoNearStation) at ($(StationFallback) + (1.6, -2)$) [text centered, fill=white, draw, rectangle, minimum width=1.5cm, text width=4.4em, font=\large]{Go near Station};
        		\draw[-latex] (StationFallback.south) -- (GoNearStation.north);
		    \end{tikzpicture}}
		\caption{Delivery Tool Tree: subtree controlling tool delivery tasks.}
		\label{fig:DeliverToolTree}
	\end{center}
\end{figure}

A hierarchical~\ac{BT}, i.e., the \textit{Main Tree}, encodes the behavior of each~\ac{UAV}, as depicted in Figs.~\ref{fig:MainTree},~\ref{fig:someBTs} and~\ref{fig:DeliverToolTree}. This \textit{Main Tree} takes as input the list of tasks from the \textit{High-Level Planner}.
If the mission is not over, the battery level is checked, and if the~\ac{UAV} has a wrong tool, this is dropped  (\textit{Drop Tool Tree}). Depending on the next assigned task, the corresponding subtree (i.e., \textit{Monitoring Task Tree}, \textit{Inspection Task Tree} or \textit{Delivery Tool Task Tree}) is executed, if in \textit{Idle} the~\ac{UAV} goes to the recharging station. If a failure is detected (i.e., a communication drop-out or a lack of battery), there is an emergency protocol that empties the~\ac{UAV}'s task queue, so that the~\ac{BT} detects no tasks and activates the \textit{Idle} condition to head the recharging station. Meantime, a failure to execute the task is reported to the \textit{High-Level Planner}, who triggers a replanning procedure that redistributes the tasks among the available~\acp{UAV} to ensure the mission is accomplished successfully.

\section{Simulation Results}
\label{sec:experimental_results}

\begin{figure}
	\centering
	\centering
	\begin{tikzpicture}
	
	\node[anchor=south west,inner sep=0] (img) at (0,0) { 
		\adjincludegraphics[width=0.45\textwidth, trim={{0.0\width} {0.08\height} {0.0\width} 
		{0.1\height}}, clip]{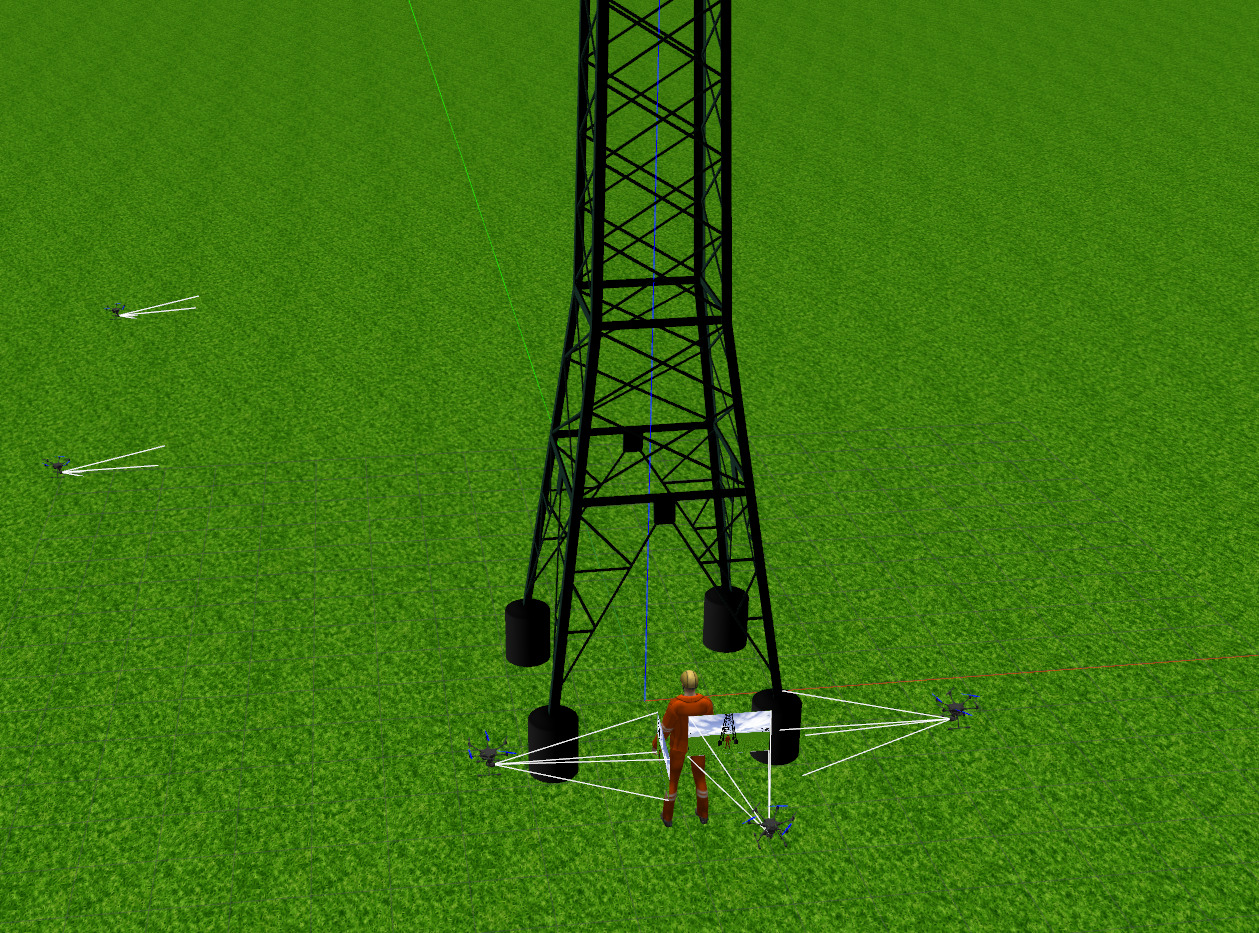}};
	\begin{scope}[x={(img.south east)},y={(img.north west)}]
	
	\draw [white, ultra thick] (0.06, 0.5) circle (0.05);
	\draw [white, ultra thick] (0.10, 0.7) circle (0.05);
	\draw [white, ultra thick] (0.4, 0.135) circle (0.05);
	\draw [white, ultra thick] (0.62, 0.04) circle (0.05);
	\draw [white, ultra thick] (0.76, 0.19) circle (0.05);
	
	\draw [white, dashed, ultra thick] (0.55, 0.145) ellipse (0.05 and 0.125);
	
	\end{scope}
	\end{tikzpicture}
	\caption{A snapshot of the \textit{monitoring} scenario. Solid and dashed circles indicate the~\acp{UAV} and the human worker, respectively.}
	\label{fig:experiments} 
\end{figure}
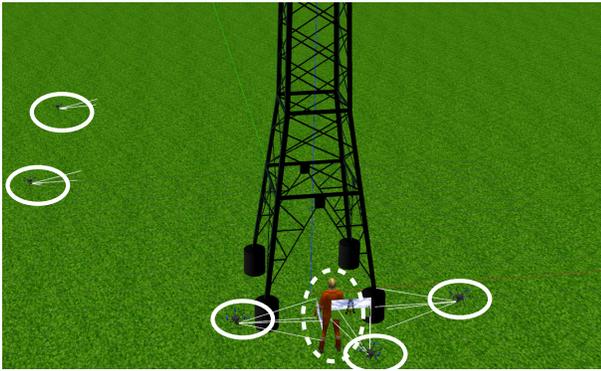

This section showcases the feasibility of the proposed software architecture through several use cases in a realistic simulation setup. In particular, we run the software architecture in a quite close to real application scenario using the Gazebo robotic simulator, exploiting the advantages of Software-In-The-Loop simulations~\cite{Silano2019SMC}. 
The software stack was coded using the~\ac{ROS} Melodic Morenia running on Ubuntu 18.04. The \textit{BehaviorTree CPP} library\footnote{\url{https://github.com/BehaviorTree/BehaviorTree.CPP}} has been used to code the \acp{BT}. All simulations were performed on a laptop with an Intel-Core i7-7700 processor ($\SI{3.60}{\giga\hertz}$) and 16GB RAM. Figure~\ref{fig:experiments} depicts a snapshot of the monitoring task, while illustrative videos with the simulations are available at~\url{http://mrs.felk.cvut.cz/bt-planner}. The code has been released as open-source\footref{footnote:softwareGitHub} making it possible to replicate the obtained results.

In order to test the interaction of the~\acp{UAV} with a realistic environment, the simulated scenario includes power towers of $\SI{20}{\meter}$ high along with the corresponding wires and a human worker. All Gazebo models and mesh files have also been made available\footref{footnote:softwareGitHub}. For the sake of simplicity and ease of experimentation, human gestures to command high-level actions and low-level controllers to perform the tasks have been simulated using console user inputs and idle states, respectively. Note that this does not imply a loss of generality as human gestures and low-level controllers are only input and output interfaces for the proposed software architecture, respectively.



Additionally, the simulation tests use a~\ac{ROS} node modeling the~\ac{UAV} battery discharge to generate unexpected events such as a sudden lack of battery. Thus, the run experiments demonstrate the applicability of the software architecture in emergency situations, where a replanning is triggered to reassign the pending tasks to the available~\acp{UAV}. In this scenario, recharging stations are modeled as waypoints to reach for the~\acp{UAV}, and then land, thus simulating a battery replacement operation.

\section{Conclusions}
\label{sec:conclusions}

This paper has presented a software architecture to support maintenance and inspection operations in power lines with a fleet of heterogeneous aerial vehicles. In particular, a two-layer software stack based on a centralized high-level planner and a distributed agent behavior manager has been proposed. The planning accounts for task priorities and~\ac{UAV} battery constraints, while the replanning allows~\acp{UAV} to react to unforeseen events during mission execution. The architecture has been designed around a set of libraries and software components that handle the interface with~\ac{UAV} low-level controllers for task execution and crew requests through human gestures. Simulations in the Gazebo robotic simulator have demonstrated the feasibility of the proposed software architecture, aiming towards the fulfillment of real-word tests. As future work, we plan to integrate an optimal layer for mission planning as well as accounting for uncertainties in task execution and workers' intentions. In addition, formal verification methods will be investigated with the aim of testing the~\ac{UAV} behaviors in as many use cases as possible.

\balance








\bibliographystyle{IEEEtran}
\bibliography{bib_short}

\end{document}